\documentclass[twocolumn]{ceurart}
\usepackage[utf8]{inputenc}

\usepackage{soul}

\usepackage{graphicx} 
\usepackage{amsmath, amssymb, dsfont}
\usepackage{physics}
\usepackage{amsfonts}
\usepackage{subcaption}
\usepackage{booktabs}
\usepackage{arydshln}
\usepackage{pifont}
\usepackage{xcolor}   
\usepackage{textcase} 
\usepackage{tabularx}
\usepackage{makecell}
\usepackage[normalem]{ulem}
\newcolumntype{Y}{>{\centering\arraybackslash}X}
\usepackage{array}
\newcolumntype{V}[1]{>{\centering\arraybackslash\rotatebox[origin=c]{90}{\makebox[#1][c]{}}}m{2cm}}
\newcolumntype{M}[1]{>{\centering\arraybackslash}m{#1}}

\usepackage{listings}
\usepackage{xcolor}
\usepackage{listings}
\usepackage{multirow}

\renewenvironment{quote}
  {\list{}{\leftmargin=10pt \rightmargin=10pt \topsep=5pt \parsep=5pt \itemsep=5pt}%
   \item\relax}
  {\endlist}

\definecolor{darkred}{rgb}{0.6, 0, 0}
\definecolor{darkgreen}{rgb}{0, 0.5, 0}

\date{March 2024}
\copyrightyear{2025}

\copyrightclause{}

\conference{}

\begin{document}

\title{An Efficient Long-Context Ranking Architecture With Calibrated LLM Distillation: Application to Person–Job Fit}

 \author[1]{Warren Jouanneau}[
 orcid=0000-0003-4973-2416,
 email=warren.jouanneau@malt.com,
 ]
 \fnmark[1]

 \author[1]{Emma Jouffroy}[
 ]
 \fnmark[1]

 \author[1]{Marc Palyart}[%
 ]
 \address[1]{Malt, 33000 Bordeaux, France}
 \fntext[1]{These authors contributed equally.}

\begin{abstract}
Finding the most relevant person for a job proposal  in real time is challenging, especially when resumes are long, structured, and multilingual. In this paper, we propose a re-ranking model based on a new generation of late cross-attention architecture, that decomposes both resumes and project briefs to efficiently handle long-context inputs with minimal computational overhead. 
To mitigate historical data biases, we use a generative large language model (LLM) as a teacher, generating fine-grained, semantically grounded supervision. This signal is distilled into our student model via an enriched distillation loss function. The resulting model produces skill-fit scores that enable consistent and interpretable person–job matching. Experiments on relevance, ranking, and calibration metrics demonstrate that our approach outperforms state-of-the-art baselines.
\end{abstract}

\maketitle

\section{Introduction}
\label{sec:intro}

The application of machine learning to Human Resources (HR) data has led to significant advances in tasks such as career path prediction~\cite{decorte2023career} or skill extraction~\cite{zhang2022skill}. Central to this domain is the challenge of matching talent to projects, a core component of modern recommender systems. On large-scale platforms, this task requires an automated process to navigate thousands of potential candidates efficiently. This challenge is especially acute in freelancing marketplaces,
like Malt, Europe's leading freelancing marketplace with over 850.000 freelancers among a broad range of industries (from technical roles like back-end development to creative fields such as design),
where precision is critical: freelancers are often expected to contribute effectively upon starting a project. 

A common approach for such systems is a two-stage pipeline consisting of a retrieval and a ranking phase. Our work focuses on the latter, where creating an effective and scalable model presents several challenges. First, freelancer profiles and project briefs are often long, structured documents written in multiple languages. While lexical matching methods fall short in capturing deep semantic meaning, many transformer-based models that excel at this are limited to short input lengths or require significant computational resources not suitable for real-time inference. Second, the performance of supervised learning-to-rank models is highly dependent on the quality of training data. Real-world recommender systems often produce sparse, biased interaction data \cite{schnabel2016recommendations}. Selection and exposure biases stem from users engaging only with visible items, making missing data ambiguous \cite{steck2010training}. Presentation and popularity biases inflate the most popular and top-ranked items, limiting diversity \cite{ge2010beyond}. In addition, interaction histories can reinforce stereotypes and underrepresent certain user groups \cite{wang2023survey}.

Beyond the challenges of training on biased data, neural ranking models present another critical limitation in HR contexts: their outputs frequently lack the global score calibration that is essential for interpretability and consistent, query-invariant comparisons. A promising strategy to instill these desired properties while maintaining efficiency is knowledge distillation \cite{hinton2015distilling}. This paradigm, which transfers the nuanced judgments of a large "teacher" model to a compact "student" model, has emerged as a powerful approach for complex reranking tasks and provides a pathway to bridge this gap.

To address all of these challenges, we propose a novel framework that distills the semantic reasoning of a generative Large Language Model (LLM) into a lightweight, efficient, and calibrated ranking model. We use the LLM as a "teacher" to generate fine-grained and semantically grounded relevance scores, creating a robust supervisory signal that overcomes the limitations of biased historical data. Our contributions are twofold:
\begin{itemize}

\item \emph{A Distillation Framework for a Calibrated and Interpretable Semantic Score}: We introduce a distillation framework to produce a relevance score that is both interpretable and semantically calibrated. First, to overcome the limitations of biased and sparse historical interaction data, our method uses a generative LLM to provide a semantically calibrated relevancy score. Second, we combine a distillation loss sensitive to both ranking and score magnitude with direct score supervision to improve calibration.
Overall, This ensures that the final score produced by our model has a consistent meaning across different freelancer-project pairs, making it suitable for ranking and for use in downstream business applications.

\item \emph{A Lightweight, Long-Context Reranking Architecture}: We propose an efficient student model that processes long-form profiles and project briefs by decomposing them into structured utterances. Its cross-attention comparison block, inspired by late interaction mechanisms, effectively captures fine-grained semantic alignment while remaining computationally inexpensive for real-time inference.

\end{itemize}

The remainder of this paper is organized as follows. Section~\ref{sec:related} refines our problem statement and reviews related work on HR recommendation systems and semantic ranking algorithms. Section~\ref{sec:method} presents in depth our dataset creation and ranking algorithm alongside the training objective. Section~\ref{sec:experiments} describes the experimental setup and results. We conclude in Section~\ref{sec:conclusion} with future directions.

\section{Related Work}
\label{sec:related}

In HR applications, ranking or re-ranking is used to refine candidate-job matches after an initial retrieval stage, aiming to surface the most relevant candidates efficiently. One approach to candidate-job ranking uses single-stream comparison models, from Recurrent Neural Networks (RNNs) with attention \cite{qin2018enhancing} to Graph Convolutional Networks (GCNs) that capture structured relationships \cite{yang2022modeling}.

Driven by real-time constraints in deployment, bipartite architectures (bi-encoders) became popular. These encode candidates and jobs separately before applying a similarity function. Early implementations relied on convolutional encoders and contrastive training \cite{zhu2018person, maheshwary2018matching}, while more recent work adopted transformer-based models such as ConsultantBERT \cite{lavi2021consultantbert}, extending Sentence-BERT \cite{sentence-bert} to the HR domain. Contrastive performance has been further improved via data augmentation—either through heuristics \cite{yu2024confit} or LLM-generated synthetic resumes \cite{yu2025confit}, and federated learning setups \cite{zhang2023fedpjf}.

To go beyond independent encoding, graph-based methods capture richer structural relationships. Some use external knowledge graphs to learn job and candidate embeddings \cite{ramanath2018towards}, while others incorporate relational structure into transformers branches using losses  \cite{zhao2021embedding} or relational GCNs trained jointly with encoders \cite{bian2020learning}.

Recent attention-based comparison models reintroduce interaction layers, similar to earlier single-stream designs, to capture fine-grained alignments between job requirements and candidate profiles, improving ranking accuracy \cite{shao2023exploring}.

\paragraph{Large, Structured, Multilingual Documents.}
HR platforms operating across countries must match candidates to jobs using long, structured, and multilingual documents \cite{zhu2018person, lavi2021consultantbert, yu2024confit}. To capture structural information, such as education, skills, and experience, some approaches adopt hierarchical or field-aware encoders that reflect document layout \cite{qin2018enhancing, zhu2018person}, while others use segment-level encoding and aggregation to align subfields across candidate and job profiles \cite{yu2024confit, malt-retrieval}. Language alignment in bi-encoders has been tackled through distillation techniques \cite{reimers2020making}, as in multilingual Sentence-BERT \cite{sentence-bert}. Arctic-Embed-v2 \cite{yu2024arctic} extends this paradigm to both long-context and compact variants. To incorporate interaction modeling without retraining, late interaction mechanisms such as ColBERT \cite{khattab2020colbert} have been introduced. Alternatively, single-stream rerankers such as Qwen3 \cite{qwen3embedding} also support long multilingual inputs.

\paragraph{Generative Large Language Models for Reranking.}
Generative large language models (LLMs) have recently emerged as competitive zero-shot rerankers, for large multilingual documents, by leveraging their capacity to reason over document-query pairs in natural language. Unlike dense retrieval or cross-encoders trained on annotated datasets, generative LLMs can predict ranking permutations directly \cite{ma2023zero}, or assign relevance scores to each candidate \cite{zhuang2024setwise} without needing further training. Several techniques have been proposed to improve reranking performance: prompt design algorithms \cite{jin2025apeer} and output manipulation strategies \cite{gao2025llm4rerank}. In-context reranking (ICR) methods use the LLM’s attention dynamics to infer preferences across candidates \cite{chen2024attention}. Others explore using internal representations such as first-token embeddings to train lightweight rerankers on top of frozen LLMs \cite{reddy2024first}. These approaches offer a flexible, instruction-following alternative to traditional supervised rankers, especially in zero-shot setups where labeled training data is unavailable or biased. However, their high latency and computational cost are often not compatible with live inference.

\paragraph{Distillation from Generative LLMs.}

To address the impracticality of using directly generative LLMs in production, they can serve as teachers in a distillation setup—generating either synthetic training data (e.g., queries or candidate items \cite{yu2025confit}) or soft supervision signals (e.g., scores or ranking permutations). This information is then used to train smaller, efficient student models. For example, \citet{sun2023chatgptrerank} demonstrate that listwise permutations produced by LLMs can be treated as ground-truth orderings for training student models using RankNet~\cite{burges2005ranknet}, outperforming traditional supervised baselines. Other works, such as \citet{shang2025marginnogt}, explore score-level distillation by fine-tuning on margin-aware objectives tailored for LLM-generated supervision. Their approach builds on margin-MSE~\cite{hofstatter2020marginmse}, adapting it for better transfer from LLM score distributions. Although these methods might bring biases from their own training data\cite{wang2024large}, for example, minor group favoritism or skewed token priors, they highlight the promise of generative LLMs as a rich source of ranking signal to tackle many of the biases present in recommender training data.

\paragraph{Ranking Calibration.}
In ranking systems, score calibration is critical when downstream tasks rely not just on item orderings but also on the absolute values of predicted scores—for example, in multi-stage retrieval, risk-aware decisions, or interpretability of results. However, most ranking losses do not enforce calibration, often producing scores that lack a consistent scale \cite{yan2022scale}. Solutions include modeling uncertainty via dropout or ensembles \cite{penha2021calibration}, leveraging LLM-generated explanations as auxiliary signals \cite{yu2024explain}, or applying post-hoc corrections like binning. Some approaches directly integrate calibration into the loss function, including recent work on distillation-aware objectives such as CLID \cite{gui2024clid}. Together, these methods help ensure that ranking models produce scores that are meaningful, comparable across queries, and robust for real-world applications.

\section{Approach}
\label{sec:method}

Let $\mathbb{P}$ and $\mathbb{F}$ denote the sets of all projects and freelancers, respectively. Each project $p \in \mathbb{P}$ is represented by a brief document $x_p$, and each freelancer $f \in \mathbb{F}$ by a profile $x_f$. These documents are composed of structured textual sections denoted $s_{d, l}$, where $d \in \{\mathbb{P}, \mathbb{F}\}$ and $l \in L_d$ refers to a document-specific section type. Thus, each document is defined as:
\[
x_d = \{s_{d, l} \mid l \in L_d \}.
\]

Our goal is to obtain a model $M_\theta$ with parameters $\theta$ that estimates a continuous skill fit relevance score $s_{f,p} \in [0,1]$ for any pair $(f, p)$:
\[
M_\theta(x_f, x_p) = s_{f,p}.
\]

This score is intended to reflect the skill-fit between a freelancer and a project, enabling skills based ranking such that:
\[
\forall f, f' \in \mathbb{F},\quad s_{f,p} > s_{f',p} \Rightarrow f \succ_p f'.
\]

Beyond ranking accuracy, we aim for the scores to be interpretable and comparable across different project–freelancer pairs, i.e., semantically calibrated. This enables to use them as meaningful features for downstream applications such as explanation, policy decisions, or business-level ranking.

Let $\mathcal{R}$ be a fixed reference set of matching interpretations (e.g., \textit{“unqualified,” “partial match,” “strong match”}, ...), and let $\phi: [0,1] \rightarrow \mathcal{R}$ be a mapping from scores to interpretation. In this setting, semantic calibration requires:
\begin{equation}
\label{eq:semantic-calibration}
    \forall_{\substack{f, f' \in \mathbb{F} \\ p, p' \in \mathbb{P}}} \,
    \left(s_{f,p} = s_{f',p'}\right) \Rightarrow 
    \left(\phi(s_{f,p}) = \phi(s_{f',p'})\right).
\end{equation}

This ensures that identical scores carry consistent meaning across contexts, facilitating explanations and cross-project comparability. \\

\begin{figure}
    \centering
    \includegraphics[width=\linewidth]{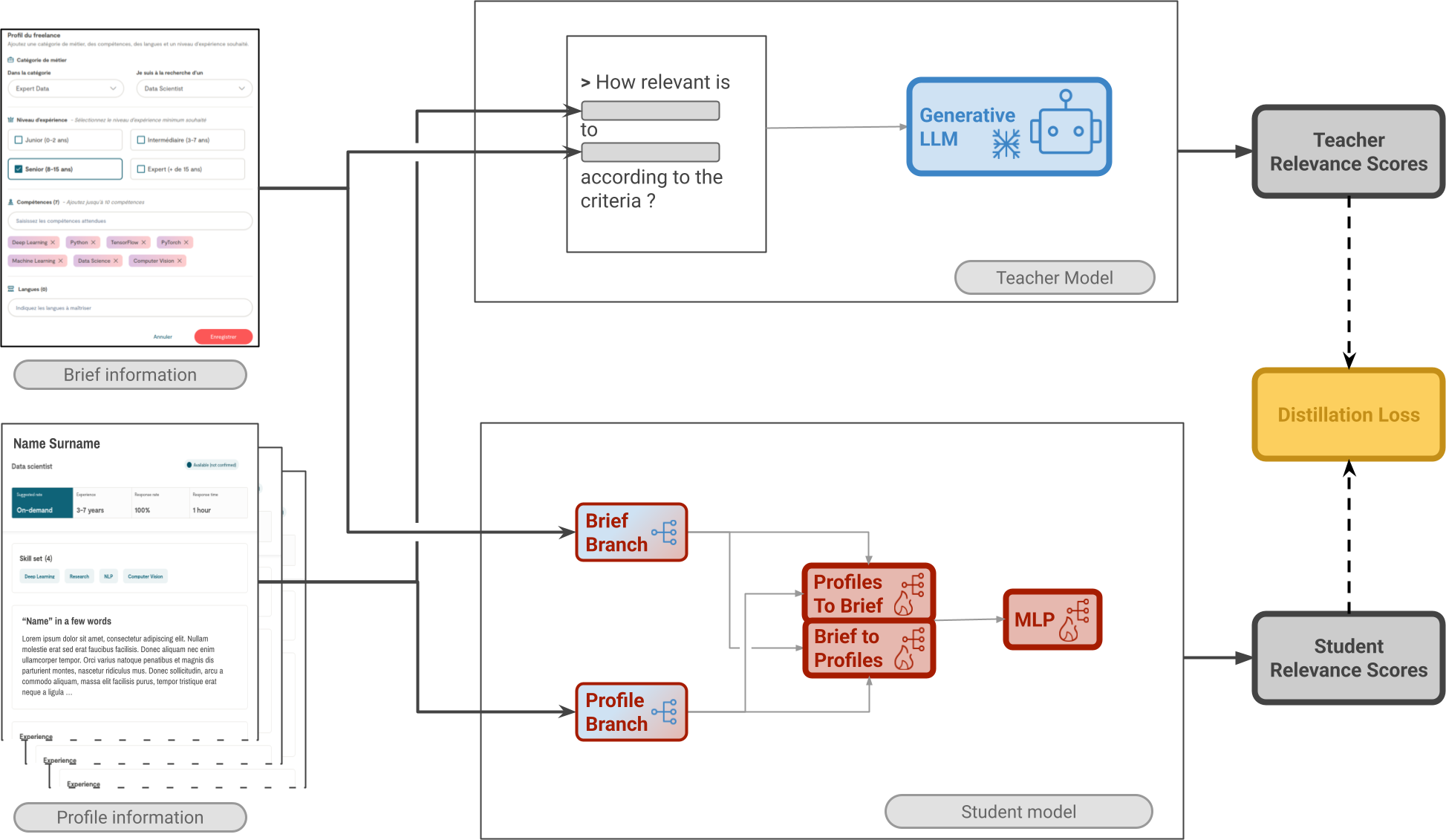}
    \caption{Overview of the semantic score distillation pipeline: a generative LLM assigns semantic labels that are mapped to reference scores to form a calibrated teacher model $M_{\text{teacher}}$, whose outputs are used to train a lightweight student model $M_{\text{student}}$.}
    \label{fig:distillation}
\end{figure}

To obtain such calibrated scores, we assume the existence of a model $M_{\theta'}$ that assigns a relevance category $r \in \mathcal{R}$ to any freelancer–project pair $(x_p, x_f)$:
\[
M_{\theta'}(x_p, x_f) = r.
\]

Conceptually, an approximate inverse mapping $\tilde{\phi}^{-1}: \mathcal{R} \rightarrow [0,1]$, can be constructed to associate each semantic category with a reference score:
\[
\tilde{\phi}^{-1}(r) = \mathbb{E}_{s \sim p(s \mid \phi(s) = r)}[s].
\]

In results, a teacher model $M_{\text{teacher}}$ can be defined as :
\[
M_{\text{teacher}}(x_p, x_f) = \tilde{\phi}^{-1}(M_{\theta'}(x_p, x_f)),
\]
which outputs semantically calibrated scores grounded in the interpretation space $\mathcal{R}$.

To make inference scalable, we train a compact student model $M_{\text{student}}$ that mimics $M_{\text{teacher}}$:
\begin{equation}
\label{eq:studapproxteach}
M_{\text{student}}(x_p, x_f) \approx M_{\text{teacher}}(x_p, x_f).    
\end{equation}

This distillation process, illustrated in Figure~\ref{fig:distillation}, enables us to preserve the hypothetic semantic calibration and interpretability of teacher-generated scores in a lightweight model suitable for real-time deployment.

\subsection{Generative LLM relevance scoring} 
\label{gen_ai_distillation}

Having outlined our general approach and the necessity of semantic calibration, we now detail our methodology for obtaining calibrated training data. \\

In the absence of high-quality, semantically calibrated data, we propose leveraging generative large language models (LLMs) as teachers to generate skill-fit supervision signals. Indeed, as stated in the introduction (Section \ref{sec:intro}), our historical data is mostly sparse, noisy, biased and uncalibrated. In contrast, a generative LLM can evaluate tasks within a defined context, allowing us to create more fine-grained and semantically accurate supervision signals, denoted as $s^t_{f,p} \in [0,1]$. We assume that a generative LLM possesses sufficient semantic reasoning capacity to determine the correct relevance category $r \in \mathcal{R}$ and execute the inverse mapping $\tilde{\phi}^{-1}$.\\

To generate these scores, the following context is defined in the model's prompt:

\begin{quote}
\small\textit{
" You are an objective assistant in a freelancer-job matching platform. Given a job description and several freelancer profiles, evaluate each freelancer's suitability for the job. Provide a concise reasoning and a score between 0 and 1 for each freelancer.  "
}
\end{quote}
Within this context, the model is instructed to return both a score and a reasoning justification which has been shown to provide better results for complex tasks \cite{wei2022chain}. 
Both the job description and the freelancer profile are provided in plain text, with each section clearly delineated by specific indicators (e.g., "skills:", "description", etc.).  
By providing these information, the aim for the model is to generate a score that is aligned with the context and independent of user behavior artifacts, such as :   
\begin{equation}
\label{eq:teacherScore}
M_{\text{teacher}}(x_p, x_f) = s^t_{f,p} \in [0,1].
\end{equation}

To ensure semantic calibration, the prompt also contains a predefined matching interpretation set $\mathcal{R}$, aligning each score with a relevance category : 
\begin{lstlisting}[basicstyle=\small, caption={Reference list within the teacher model's prompt}, breakindent=14pt]
0.0: No relevant skills or experience. Completely unable to perform the job.  
0.2: Minor relevance. Few matching skills or limited experience.High chance they will be unable to perform the job.  
0.4: Moderate match. Some relevant skills or experience. Would probably not be able to do the job.  
0.6: Good match. Mostly relevant skills and experience. Can perform with some ramp-up.  
0.8: Strong match. Highly relevant skills and experience. Ready to perform well.  
1.0: Perfect match. Skills and experience fully aligned with job needs. Expert on the topic.  
\end{lstlisting}

By inserting this predefined mapping directly into the prompt design, the LLM is able to select a score that best represents the skill-fit between the freelancer and the job, according to the defined semantic categories. \\

To promote better instruction understanding and score consistency, the LLM is prompted with twelve freelancer profiles per project, with at least one  "\textit{unsuitable}" and one "\textit{perfect}". While each candidate is scored independently, batching them in the same prompt encourages the model to improve the score quality, without compromising semantic interpretability. \\

While this approach allows to obtain the expected semantically calibrated relevancy score, it remains computationally and environmentally expensive for large-scale inference. To overcome such limitations, a lightweight student model can be designed to approximate the LLM's output, as detailed in the following sections.

\subsection{Light relevance scoring architecture}
\label{proposed_architecture}

\begin{figure}
    \centering
    \includegraphics[width=\linewidth]{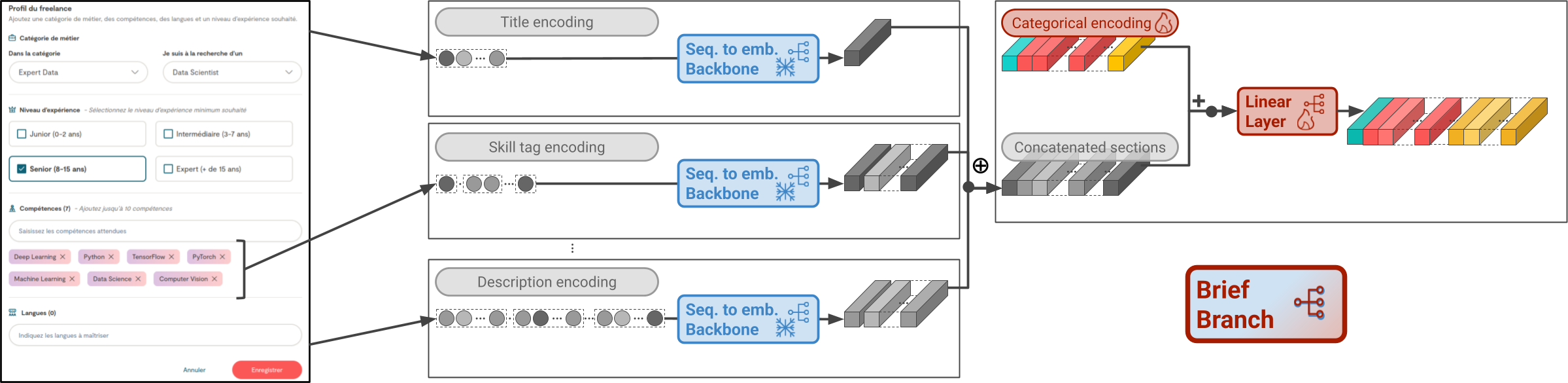} \\
    \vspace{1em}
    \hrule height 0.5pt
    \vspace{1em}
    \includegraphics[width=\linewidth]{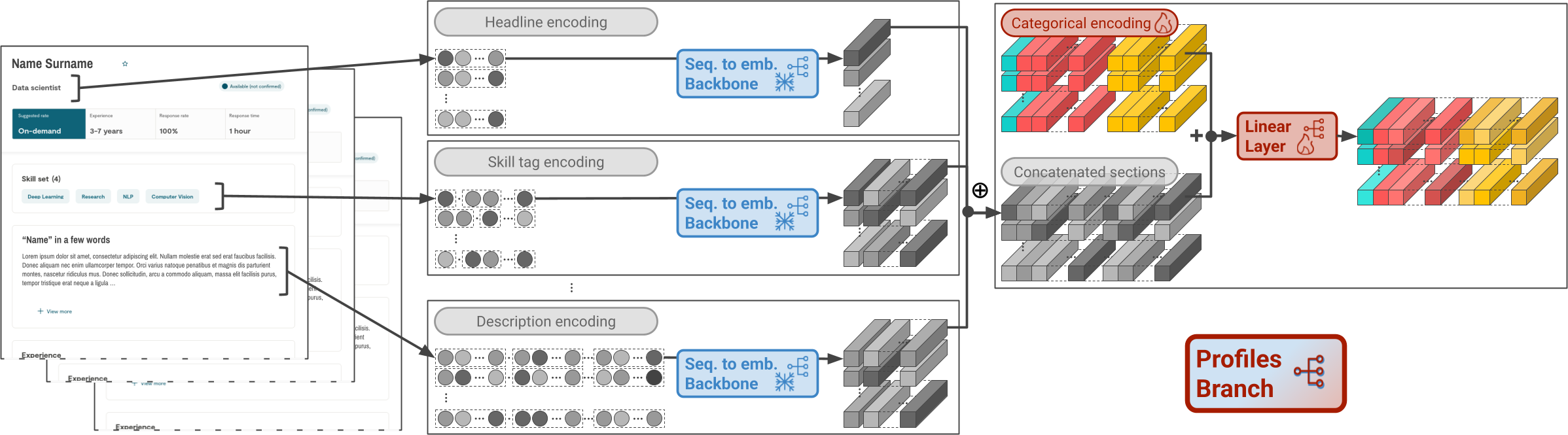} \\
    \vspace{1em}
    \caption{\textbf{Illustration of the proposed brief encoder branch (top) and profile encoder branch (bottom).} Similar sections across freelancer profiles are first projected into a latent space (shown in the first three blocks). Positional encodings are then added, and the resulting embeddings are combined with learned section-type embeddings to retain structural information. These enriched vectors are finally passed through a linear layer.}
    \label{fig:encoder-architecture}
\end{figure}

\begin{figure*}
    \centering
    \includegraphics[width=0.8\linewidth]{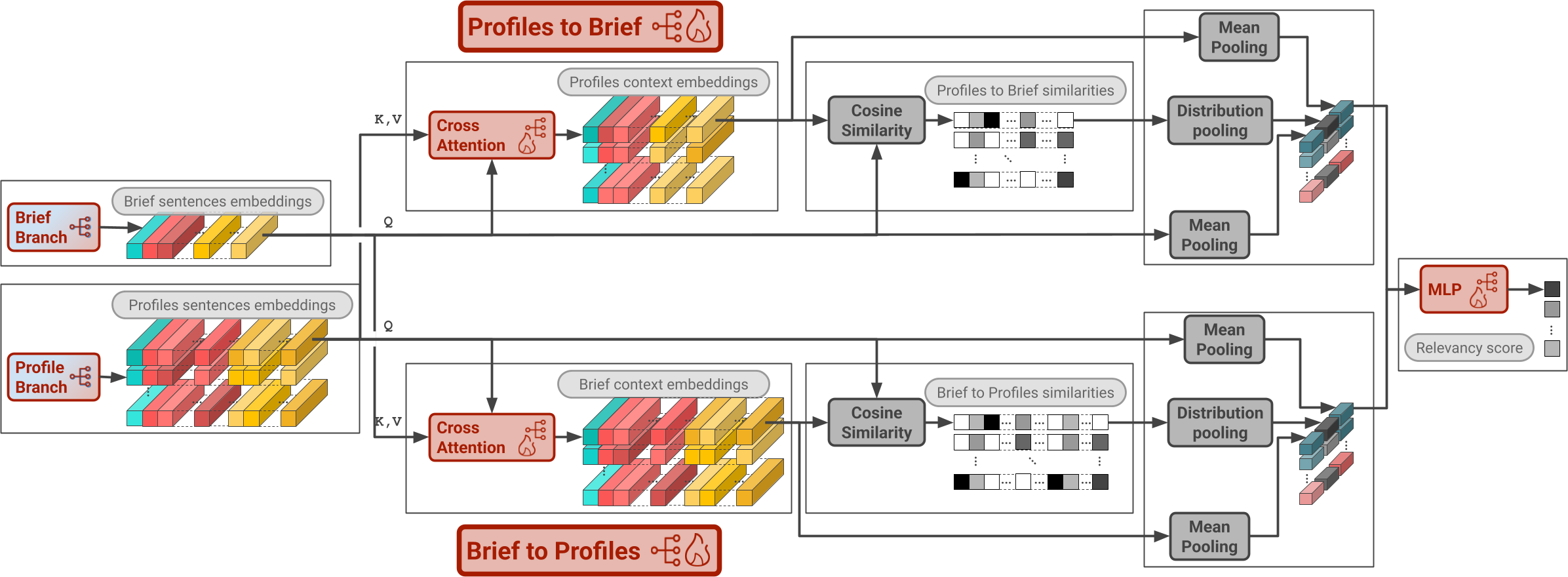} 
    \caption{\textbf{Overview of the proposed comparison architecture.} The first two blocks (“Profiles to Brief” and “Brief to Profiles”) depict the multi-head cross-attention mechanism. The following stages show the computation of cosine similarity distributions, statistical pooling into fixed-size features, and final concatenation before scoring via an MLP.}
    \label{fig:comparison-architecture}
\end{figure*}

To retain the semantic abilities of the teacher model within a more compact one, the proposed architecture for the student model, $M_{\text{student}}$, illustrated in Figure~\ref{fig:distillation}, is composed of two main components:

\begin{enumerate}
    \item \textbf{Document Encoders}: One encoder branch for each document type (i.e., project briefs and freelancer profiles), as shown in Figure~\ref{fig:encoder-architecture}. 
    \item \textbf{Comparison Block}: An attention-based comparison module, detailed in Figure~\ref{fig:comparison-architecture}.
\end{enumerate}

The following sections describe each of these components in detail.

\subsubsection{Leveraging Pre-trained Multilingual Sentence Encoder — Document Encoders}

To encode documents, we build upon our previous work on retrieval models~\cite{malt-retrieval} within the same project-freelancer matching setting. Each document $x_d$ (either a brief or a profile) is encoded independently by processing its structured textual sections $s_{d,l}$ using a pre-trained multilingual sentence encoder and a categorical encoding. This process is illustrated in Figure~\ref{fig:encoder-architecture}. \\

Unlike our previous token-level model, each section $s_{d,l}$ is segmented into minimal textual units that are referred to as \textbf{utterances}
\[
s_{d,l} = \{u_{d,l,i} \mid i = 1 \cdots n_{d,l} \}.
\]
This strategy aligns with the intended use of \textit{sentence-BERT} models~\cite{sentence-bert} and the short length of utterances simplifies encoding, enabling the use of smaller backbones. 

Utterances are defined differently depending on the section type:
\begin{itemize}
    \item For paragraph-based sections (e.g., descriptions), utterances correspond to individual sentences.
    \item For tag-based sections, each tag is treated as a separate utterance.
    \item Titles are encoded as single-utterance sequences.
\end{itemize}

Each utterance $u_{d,l,i}$ is processed by the pre-trained sentence encoder backbone and enriched with a learned categorical encoding $\vb*{e}_{\text{categorical}_l}$ specific to section type $l$. The resulting vector is passed through a linear layer:
\begin{equation}
\vb*{e}_{d,l,i} = W_d \cdot \left( \operatorname{Backbone}(u_{d,l,i}) + \vb*{e}_{\text{categorical}_{l}} \right) + \vb*{b}_d.
\end{equation}
This categorical encoding helps preserve structural information. While the sentence encoder backbone remains frozen, both the categorical encoding and the projection layer are trained, effectively adapting the general-purpose encoder to our domain-specific skill matching task.

Thus, each document ($x_p$ or $x_f$), is ultimately represented by a sequence of utterance embeddings:
\begin{equation}
\begin{split}
    &E_f = \{\vb*{e}_{f,i} \mid i = 1 \cdots \sum_{l \in L_f} n_{f,l} \}, \\
    &E_p = \{\vb*{e}_{p,i} \mid i = 1 \cdots \sum_{l \in L_p} n_{p,l} \},
\end{split}
\end{equation}
where $n_{f,l}$ and $n_{p,l}$ denote the number of utterances per section and document.\\

This utterance-based encoding strategy significantly reduces the computational cost of processing long documents. Since most of the computational burden lies in encoding, the backbone's utterance embeddings can be pre-computed and cached to accelerate training. In production, $E_f$ can be computed and stored in advance, leaving only the project's utterances to be encoded at inference time, along with the final comparison block, presented in the next section.

\subsubsection{From Two Sequences of Embeddings to a Similarity Distribution — Comparison Block}

Inspired by the late interaction mechanism~\cite{khattab2020colbert}, similarities between the obtained embedding sequences are computed  to compare briefs and profiles. However, instead of computing only the maximum similarity per brief embedding across profile embeddings, a two-step approach that better models mutual interest is adopted. Indeed, we hypothesize that this process can capture more complex interactions (illustrated in Figure~\ref{fig:comparison-architecture}).

First, context-aware embeddings ${E_{\text{context}}}_f$ and ${E_{\text{context}}}_p$ are derived using cross-attention. Then, similarity distributions $\mathcal{S}_{p,f}$ and $\mathcal{S}_{f,p}$ are computed between the original embeddings and their respective context vectors. \\

To compute ${E_{\text{context}}}_f$, multi-head attention from the brief to the profile embeddings {is applied:
\begin{equation}
\begin{aligned}
\operatorname{MultiHead}(E_p, E_f, E_f) &= \operatorname{head}_1 \,\Vert\, \cdots \,\Vert\, \operatorname{head}_k \\
&= E_{\text{context}_f},
\end{aligned}
\end{equation}
where each attention head is defined as:
\begin{equation}
\operatorname{head}_i = \operatorname{softmax} \left( \frac{E_p W_i^Q (E_f W_i^K)^\top}{\sqrt{d}} \right) E_f W_i^V.
\end{equation}
This results in a new sequence ${E_{\text{context}}}_f$ of the same length as $E_p$, where each embedding $\vb*{e}_{\text{context}_{f,i}}$ reflects the best-matching combination of profile content for the corresponding brief utterance $\vb*{e}_{p,i}$. \\

To assess how well this profile context aligns with the original brief, their pairwise cosine similarities are computed:
\begin{equation}
\small
\mathcal{S}_{p,f} = \left\{ \operatorname{sim}(\vb*{e}_{p,i}, \vb*{e}_{\text{context}_{f,i}}) \;\middle|\; i = 1 \cdots \sum_{l \in L_p} n_{p,l} \right\}.
\end{equation}
This results in a similarity distribution with one score per brief utterance. 

Symmetrically, the process is reversed to account for mutual interest. First, the profile is attended to the brief :
\begin{equation}
\operatorname{MultiHead}(E_f, E_p, E_p) = E_{\text{context}_p},
\end{equation}
then the corresponding similarity distribution is computed:
\begin{equation}
\small
\mathcal{S}_{f,p} = \left\{ \operatorname{sim}(\vb*{e}_{f,i}, \vb*{e}_{\text{context}_{p,i}}) \;\middle|\; i = 1 \cdots \sum_{l \in L_f} n_{f,l} \right\}.
\end{equation}

These distributions, $\mathcal{S}_{p,f}$ and $\mathcal{S}_{f,p}$, provide a detailed view of skill alignment: how well a profile matches the most relevant parts of a brief, and vice versa. They form the basis for the final scoring step, described in the following section.

\subsubsection{Distribution pooling and scoring}

Since the similarity distributions $\mathcal{S}_{p,f}$ and $\mathcal{S}_{f,p}$ vary in length, $\sum_{l \in L_{p}} n_{p,l}$ and $\sum_{l \in L_{f}} n_{f,l}$ respectively, the original late interaction mechanism aggregates them using only a sum over brief-wise similarities to produce a score.

Instead, more expressive statistical pooling operations are computed over both distributions to produce fixed-size feature vectors more suitable for scoring. Specifically, we extract descriptive statistics from both $\mathcal{S}_{p,f}$ and $\mathcal{S}_{f,p}$, defined as 
\[
\operatorname{desc}(\mathcal{S}) = 
\bigl[\min(\mathcal{S}), \max(\mathcal{S}), \mu(\mathcal{S}), \sigma(\mathcal{S}), \gamma_1(\mathcal{S}), \gamma_2(\mathcal{S})\bigr],
\]
where $\mu$ denotes the mean, $\sigma$ the standard deviation, $\gamma_1$ the skewness, and $\gamma_2$ the kurtosis 
(see Appendix~\ref{sec:stats-interaction}).
These richer features are hypothesized to better capture the interaction dynamics between the documents' utterances. \\

Finally, these descriptive statistics are concatenated with the averaged pooled embeddings from the two branches, $\overline{\vphantom{E^{e}}E}_p$ and $\overline{\vphantom{E^{e}}E}_f$, as well as the averaged context embeddings ${\overline{\vphantom{E^{e}}E}_\text{context}}_f$ and ${\overline{\vphantom{E^{e}}E}_\text{context}}_p$, forming the input to a multi-layer perceptron (MLP):
\begin{equation}
\label{eq:student}
    \operatorname{MLP} \left(
    \begin{aligned}
        & \operatorname{desc}(\mathcal{S}_{p,f})
        \,\Vert\, \overline{\vphantom{E^{e}}E}_p
        \,\Vert\, {\overline{\vphantom{E^{e}}E}_\text{context}}_{f} \\
        & \,\Vert\, \operatorname{desc}(\mathcal{S}_{f,p})
        \,\Vert\, \overline{\vphantom{E^{e}}E}_f
        \,\Vert\, {\overline{\vphantom{E^{e}}E}_\text{context}}_{p}  
    \end{aligned}
    \right) = s^s_{p,f}.
\end{equation}

Adding the averaged pooled embeddings enriches the similarity distributions with contextual information from the documents. \\

Only the projection layers of the branches, the two multi-head attention modules, and the MLP require training. These components are trained to approximate the semantically calibrated scores produced by the LLM teacher model via distillation, as described in the next section.

\subsection{Training objective}
\label{training_objective}

It is often possible to derive a binary relevance label from historical interaction data. We denote this indicator function as:
\[
\delta_{p,f} =
\begin{cases}
1 & \text{if freelancer } f \text{ is relevant to project } p, \\
0 & \text{otherwise}.
\end{cases}
\]
Let $\mathbb{I}$ be the set of all project–freelancer pairs for which such a label is available:
\[
\mathbb{I} = \{ (p,f) \in \mathbb{P} \times \mathbb{F} \mid \exists! \delta_{p,f} \},
\]
and let \( I \subset \mathbb{I} \) be the subset used for training.\\

When using these historical relevance labels, a standard approach is to frame the problem as binary classification. Hence, a model $M_\theta(p,f)$ can be trained using common classification losses such as Binary Cross-Entropy, Focal Loss, or Asymmetric Loss~\cite{ridnik2021asl}. This setting has been extensively used in recommender systems and information retrieval tasks~\cite{ailon2007classification, liu2009irbook}.  \\

In contrast, ranking distillation considers supervision from a teacher model that outputs continuous-valued scores rather than binary labels. One straightforward approach is to treat the teacher's score (cf. eq.\ref{eq:teacherScore}) as a regression target for the student model score, (cf. eq. \ref{eq:student}), which can be optimized using a mean squared error (MSE) objective:
\begin{equation}
    \mathcal{L}_{\text{MSE}}(I) = \frac{1}{|I|} \sum_{(p,f) \in I} \left(s^t_{p,f} - s^s_{p,f}\right)^2.
\label{eq:mse}
\end{equation}

To validate the use of teacher model scores as ground truth, we compared relevancy metrics across three settings: the teacher model, the student model trained on historical data, and the same student further trained on teacher-generated scores.

\begin{table}[ht]
\caption{
  Evaluation of different models on historical relevance labels. "Ours (Historical labels)" is trained with ASL~\cite{ridnik2021asl} on binary labels; "Ours (Gemini scores)" is trained with MSE on the teacher scores. Re. = Recall, Spe. = Specificity, R–P = R-Precision, $\Bar{\text{R}}$-$\mathcal{O}$ = Non relevant - false omission rate, see section \ref{sec:evaluation-metrics} for metric details.
  }
  \centering
  \small
  \setlength{\tabcolsep}{8pt} 
  \begin{tabularx}{\columnwidth}{@{}XX YYYY@{}}
    \toprule
    \textbf{Model} & \textbf{Training Data}
    & Re. & Spe. & R-P & $\Bar{\text{R}}$-$\mathcal{O}$ \\
    \midrule
    Gemini & \textasciitilde
    & 0.913 & 0.287 & 0.734 & 0.216\\  
    \midrule
    Ours & Historical Labels &
    0.984 & 0.058 & 0.811 & 0.413 \\
    Ours & Gemini Scores &
    0.912 & 0.208 & 0.808 & 0.401 \\
    \bottomrule
    \label{tab:gemini_vs_historical}
  \end{tabularx}

\end{table}
\vspace{-1em}

As shown in Table~\ref{tab:gemini_vs_historical}, knowledge distillation from \textit{Gemini-2.0-flash} (cf. section \ref{sec:experiments} for details) improves the model’s ability to reject non-relevant candidates compared to training on historical binary labels. While this comes with a slight decrease in recall, the ranking quality remains comparable, suggesting that supervision via soft scores enhances discriminative capacity without sacrificing relevance. Furthermore, a qualitative evaluation, based on expert curation, supported the relevance and consistency of the generated scores. 
Overall, this setup provides access to finer-grained signals, such as ranking quality and model interpretability, that are not directly measurable from historical binary labels alone. \\

However, the MSE formulation treats each $(p,f)$ pair independently, ignoring the relative ordering between candidates, which is central to ranking tasks. This \textit{point-wise} distillation approach may therefore be suboptimal for ranking supervision~\cite{liu2009irbook}. In the following sections, we explore alternative pair-wise and list-wise objectives that better align with the ranking nature of the problem.

\subsubsection{Pair-wise Distillation}

To overcome the limitations of treating interactions independently, the relative ordering between two freelancers \( f \) and \( f' \) competing for the same project \( p \) can be distilled.\\
Specifically, the score difference given by the teacher model:
\[
\Delta^t_{p, f, f'} = s^t_{p, f} - s^t_{p, f'},
\]
is used to train the student model to replicate this margin:
\[
\Delta^s_{p, f, f'} = s^s_{p, f} - s^s_{p, f'}.
\]

This approach can be implemented using the Margin MSE loss~\cite{hofstatter2020marginmse}, which compares the predicted differences between relevant and non-relevant candidates:
\begin{equation}
\begin{aligned}
    \mathcal{L}_\text{margin\_mse}(I) &= 
    \frac{1}{n} \sum_{\substack{
        (p,f) \in I \\
        \delta_{p,f} = 1 
    }}
    \sum_{\substack{
        (p',f') \in I \\
        \delta_{p',f'} = 0 \\
        p = p'
    }} 
    \left(\Delta^t_{p, f, f'} - \Delta^s_{p, f, f'}\right)^2 \\
    &= 
    \frac{1}{n} \sum_{\substack{
        (p,f) \in I \\
        (p,f') \in I \\
    }}
    \mathds{1}_{\left[f \succ_p f'\right]}
    \left(\Delta^t_{p, f, f'} - \Delta^s_{p, f, f'}\right)^2.
\label{eq:original-margin-mse}
\end{aligned}
\end{equation}
In the later, the indicator function ensures that only pairs where \( f \) is relevant and \( f' \) is not (for the same project) are included, and n is the number of such pairs. This formulation encourages the student model to preserve the relative ordering \( f \succ f' \) induced by the ground truth.

An extension~\cite{shang2025marginnogt} was proposed relaxing the dependency on ground-truth labels by computing pairwise differences over all possible interactions $(p,f),(p,f')$ for the same project:
\begin{equation}
    \mathcal{L}_\text{margin\_mse}(I) = 
    \frac{1}{n} \sum_{\substack{
        (p,f) \in I \\
        (p,f') \in I 
    }}
    \left(\Delta^t_{p, f, f'} - \Delta^s_{p, f, f'}\right)^2.
\label{eq:walmart-margin-mse}
\end{equation}
This loss encourages the student model to preserve both the ordering and the magnitude differences between the teacher’s predictions. However, it does not enforce alignment of the absolute score values themselves.

To address this, we propose coupling the margin-based objective with a pointwise MSE regression loss (Eq.~\ref{eq:mse}) computed on the teacher’s scores. This yields our combined loss:
\begin{equation}
    \mathcal{L}_\text{CMMD}(I) =
    \mathcal{L}_\text{margin\_mse}(I)
    + \mathcal{L}_\text{MSE}(I)\\
    ,
\label{eq:ours-margin-mse}
\end{equation}
which we refer to as the Calibrated Margin MSE Distillation (CMMD) loss. Empirically, this combination yields improved performance by aligning both relative and absolute semantics of the teacher's signal

\subsubsection{List-wise Distillation}

Another training strategy employs a more natural \textit{list-wise} objective. List-wise losses are well-aligned with ranking problems which aim to order a set of candidates for a given project. Early methods such as ListNet~\cite{cao2007listnet} and ListMLE~\cite{xia2008listmle} can be adapted for distillation by using teacher scores to construct ground-truth permutations. More recently, the Calibrated List-Wise Distillation (CLID), method~\cite{gui2024clid} was introduced to facilitate calibated distillation using a list-wise approach.

In the CLID framework, the scores from both teacher and student models are normalized across the candidate set \( I \) (i.e. all freelancer profiles associated with a given project \( p \)):
\[
    \hat{s}^t_{p,f} = \frac{
        s^t_{p,f}
    }{
        \sum_{(p,f') \in I}
        s^t_{p,f'}
    }
\quad \text{and} \quad
    \hat{s}^s_{p,f} = \frac{
        s^s_{p,f}
    }{
        \sum_{(p,f') \in I}
        s^s_{p,f'}
    }.
\]
A cross-entropy loss is then applied to align these two score distributions:
\begin{equation}
    \mathcal{L}_{\text{CLID}}(I) = - \frac{1}{|I|} \sum_{(p,f) \in I} \hat{s}^t_{p,f} \log(\hat{s}^s_{p,f})
\label{eq:clid}
\end{equation}

CLID can be interpreted as aligning the probabilities of each freelancer $f$ being ranked above all others within the candidate set $I$. These probabilities are derived from the normalized scores:
\[
\mathbb{P}(f \succ^t_p \{f' \}) = \hat{s}^t_{p,f}
\quad \text{and} \quad
\mathbb{P}(f \succ^s_p \{f' \}) = \hat{s}^s_{p,f}.
\]

Alternative normalization strategies (such as computing the probability of outranking only lower-scored candidates, i.e., \( \mathbb{P}(f \succ_p \{f' \mid s^t_{p,f} > s^t_{p,f'}\}) \), as in ListMLE~\cite{xia2008listmle}) proved empirically less effective. In contrast, the original normalization proposed in~\cite{gui2024clid}, which considers the full candidate set, consistently yielded better performance. \\

\section{Experiment}
\label{sec:experiments}

The following section presents our experiments, which assess the effectiveness of the proposed distillation strategy and student model architecture in generating semantically calibrated similarity scores between freelancers and project briefs suitable for large-scale deployment 

\subsection{Implementation and baselines} \label{models_description}
First, we introduce the teacher model used to generate the semantically calibrated ground-truth scores, before defining our student's model settings. 
Then, we compare our approach against a state-of-the-art re-ranking model and a small generative language model. For both baselines, we present results using publicly available pre-trained checkpoints, with and without fine-tuning for our specific use case.

\paragraph{Teacher model: Gemini 2.0.}
As a teacher model, we employed the generative LLM \textit{Gemini-2.0-flash}, supporting up to 1M input tokens. It provides structured responses with reduced latency, balancing performance and computational efficiency compared to Gemini-1.5 and Gemini-2.0-pro. Additionally, Gemini-2.0-flash offers multilingual capabilities.

\paragraph{Student model: our model.}
We use the multilingual \textit{Arctic Embed}~\cite{yu2024arctic} (extra-small variant\footnote{\url{https://huggingface.co/Snowflake/snowflake-arctic-embed-xs}}) as a shared encoder backbone in both branches. This lightweight model was chosen to efficiently handle short utterances with minimal performance tradeoff. Each branch includes a linear projection to a 32-dimensional latent space. The comparison block consists of a single 8-head multi-head attention layer. The MLP has layers of size 256, 128, 256, and 1, with GELU activations, 0.4 dropout, and no activation in the final layer (which empirically aids distillation). The architecture totals \textbf{45M parameters}, with only \textbf{135K} trainable.

\paragraph{Reranking baseline: Qwen3.}
As a strong re-ranking baseline, we evaluate \textit{Qwen3}~\cite{qwen3embedding}, using the pre-trained \textit{Qwen3-0.6B} checkpoint\footnote{\url{https://huggingface.co/Qwen/Qwen3-Reranker-0.6B}}.
This model has \textbf{596 million parameters} and supports inputs up to 32K tokens. We chose it due to its strong multilingual capabilities and its state-of-the-art performance on various re-ranking tasks. This makes it well suited for document-level semantic comparison.

\paragraph{Small generative baseline: Gemma3.}
We include the \textbf{1B-parameter} \textit{Gemma3} model~\cite{gemma3technicalreport}\footnote{\url{https://huggingface.co/google/gemma-3-1b-it}} in our evaluation. Gemma is a smaller and open-source generative LLM. Its support for multilingual inputs and long contexts (up to 128K tokens) makes it a practical and accessible alternative for approximating Gemini-style supervision and annotation quality.

\subsection{Dataset}
\label{dataset_description}
Our corpus includes project briefs created between January 1, 2023 and April 15, 2024, along with historical versions of freelancer profiles that either applied or were rejected due to lacking skills. Profile representations are recomputed to reflect their state at the time of interaction.

For evaluation, we reserve projects from January 1 to May 1, 2024, yielding a test set of \textbf{85K} interactions between \textbf{8K} projects and \textbf{78K} profile versions. The training set contains \textbf{585K} interactions across \textbf{55K} projects and \textbf{520K} profile versions. \\

To mitigate \textit{presentation bias} from training solely on historical interactions, we augment the dataset with two types of negative examples. For \textit{average} matches, additional freelancers are scored with \textit{Gemini 2.0} and those with a score between 0.4 and 0.6 are retained, adding \textbf{265K} interactions involving \textbf{256K} additional profiles. For \textit{unsuitable} matches, profiles not having any job category in common with the project are randomly sampled at batch time. \\

All supervision signals, both for training and evaluation, are derived from \textit{Gemini 2.0}, as described in Section~\ref{gen_ai_distillation}. These scores are used as ground-truth relevance labels in our experiments.

\subsection{Training Settings}
\label{training_description}
\paragraph{Our Architecture.}
We train our architecture using three supervision strategies: point-wise, pair-wise, and list-wise. For clarity, we report only the best-performing objective for each.

\textbf{Point-wise} training uses the standard mean squared error loss (MSE), as defined in Eq.~\ref{eq:mse}, denoted $\mathcal{L}_\text{MSE}$.

\textbf{Pair-wise} training employs our combined margin-based distillation loss $\mathcal{L}_\text{CMMD}$ (Eq.~\ref{eq:ours-margin-mse}), which outperforms previous formulations (Eq.~\ref{eq:original-margin-mse}, Eq.~\ref{eq:walmart-margin-mse}).

\textbf{List-wise} training uses a combination of calibration distillation loss (Eq.~\ref{eq:clid}) and $\mathcal{L}_\text{MSE}$, mentioned as $\mathcal{L}_{\substack{\text{CLID} \\ \text{+MSE}}}$ in the results section.

Point-wise batches were sampled independently. For pair-wise and list-wise, batches included one freelancer per discrete teacher score (e.g., 0.0, 0.2, …) per project, plus two synthetic \textit{unsuitable} profiles. Models were trained for 50 epochs with batches of 64 projects (~320 profiles). The frozen encoder allowed precomputing embeddings, reducing training time to ~10 hours on an NVIDIA RTX A1000. Learning rate followed linear decay starting at $0.001$:

\paragraph{\textit{Qwen3} and \textit{Gemma3}.}
Both were fine-tuned using $\mathcal{L}_\text{MSE}$ on 5,000 interactions for 10 epochs. \textit{Gemma3} was also trained with next-token prediction to match Gemini 2.0’s format. We used parameter-efficient quantized fine-tuning (e.g., QLoRA~\cite{dettmers2023qlora} with 4 bits quantization, paged AdamW 8-bit as optimizer, bf16 floating point format) to reduce compute. Training took ~1 day for \textit{Gemma3} and ~0.5 day for \textit{Qwen3} on a Tesla T4 GPU.

\textit{Qwen3} was used in a standard BERT-style cross-encoder setup, where the brief and profile were concatenated before being fed into the model. \textit{Gemma3} followed the Gemini distillation format, encoding the (freelancer, project) context and criteria set. Profiles were truncated to 2,000 tokens to fit memory, prioritizing recent relevant experience. \\

\subsection{Evaluation Metrics}
\label{sec:evaluation-metrics}
To evaluate model performances, three categories of metrics are used: (i) \textit{relevancy} metrics that evaluate the models’ ability to discriminate between relevant and non-relevant candidates; (ii) \textit{ranking} metrics, that assess the correctness of candidate ordering; and (iii) \textit{calibration} metrics that evaluate how well the predicted scores aligns with the ground-truth, indicating calibration quality.

\paragraph{Relevancy metrics.}
We define a freelancer as \textit{relevant} if $s^t > 0.5$, and \textit{non-relevant} otherwise. Based on this definition, we compute:
\begin{itemize}
    \item \textbf{Recall (Rec.)}, measuring the proportion of relevant freelancers correctly identified.
    \item \textbf{Specificity (Spec.)}, assessing the ability to correctly reject non-relevant freelancers.
    \item \textbf{R-Precision ($R$-$P$)}, which is the precision at $k$ per project, with $k$ the number of relevant freelancers.
    \item \textbf{Non-Relevant False Omission Rate ($\Bar{R}$-$\mathcal{O}$)}, an inverse analogue of $R$-$P$ that captures how many of the bottom-ranked freelancers are non-relevant.
    \item \textbf{Mean Average Precision (mAP)}, a standard metric assessing relevancy across ranks.
\end{itemize}

\paragraph{Ranking metrics.}
To evaluate the quality of the ranking itself (independent of relevance thresholds), we report:
\begin{itemize}
    \item \textbf{Mean Reciprocal Rank (MRR)}, which considers the position of the first relevant freelancer.
    \item \textbf{Normalized Discounted Cumulative Gain (NDCG)}, which accounts for the order of all relevant items, assigning higher importance to those ranked higher.
\end{itemize}

\paragraph{Calibration metrics.}
Assuming the teacher model (\textit{Gemini 2.0}) provides semantically calibrated scores, we assess how well the predicted score distributions fit the teacher's. We measure the distance between the predicted and target score distributions using:
\begin{itemize}
    \item \textbf{Mean Absolute Error (MAE)},
    \item \textbf{Difference in Means ($\Delta_\textit{mean}$)},
    \item \textbf{Difference in Interquartile Ranges ($\Delta_\textit{IQR}$)},
    \item \textbf{Wasserstein Distance}, measuring the minimal cost of transforming the predicted distribution into the ground-truth one.
\end{itemize}
Unlike the other metrics, which are normalized between 0 and 1 (with higher being better), lower values indicate better performance for these calibration metrics. \\

\subsection{Results}
\begin{table*}[ht]
  \caption{Evaluation results on the test set, including inference time. 
  Best results per metric are in \textbf{bold} and second best \underline{underlined}.}
  \centering
  \small
  \setlength{\tabcolsep}{8pt} 
  \begin{tabularx}{\linewidth}{@{}l@{}Yl *{5}{Y} *{2}{Y} *{4}{Y}@{}}
    \toprule
    \multirow{2}{*}{\makecell[l]{\textbf{Model} \\ \footnotesize(size)}}
    & \multirow{2}{*}{\makecell{\textbf{Inf.} \\ $\scriptstyle(t@1k)$}}
    & \multirow{2}{*}{\textbf{Loss}}
    & \multicolumn{5}{c}{\textbf{Relevancy} $\scriptstyle(s_t > 0.5)$}
    & \multicolumn{2}{c}{\textbf{Ranking}}
    & \multicolumn{4}{c}{\textbf{Calibration} (distances)} \\
    \cmidrule(lr){4-8} 
    \cmidrule(lr){9-10} 
    \cmidrule(lr){11-14}
     & & &
     Rec. & Spec. & R-P & $\Bar{\text{R}}$-$\mathcal{O}$ & mAP 
    & MRR & NDCG
    & MAE & $\Delta_{mean}$ & $\Delta_{IQR}$ & $W_1$ \\
    
    \midrule
    
    \multirow{2}{*}{\makecell[l]{Qwen \\ \scriptsize (0.6B)}} & \multirow{2}{*}{7.06m} &\textasciitilde 
      &  0.774 & \textbf{0.717} & \textbf{0.942} & \underline{0.552}  & 0.567     
      & 0.610  & \underline{0.973}  
      & 0.291   & \underline{0.007}    & 0.517    & 0.224      \\  
     & & $\mathcal{L}_{\text{MSE}}$
      & 0.947 & 0.214 & 0.927 & 0.456 & 0.572     
      & 0.614 & 0.970  
      & 0.145 & 0.024 & 0.071 & 0.069      \\ 
      
    \midrule
    \multirow{3}{*}{\makecell[l]{Gemma \\ \scriptsize (1B)}} & \multirow{3}{*}{\makecell[c]{ \\ 13.3m}} & \textasciitilde 
      & 0.456 & \underline{0.550} & 0.894 & 0.296 & 0.392     
      & 0.428  & 0.948  
      & 5.876   & 5.875    & 2.800    & 5.876       \\ 
     &  & $\mathcal{L}_{\text{CE}}$
      & 0.862  & 0.487 & 0.917 & 0.450 & 0.425    
      & 0.438  & 0.958  
      & 0.184   & 0.043    &  \textbf{0.000}    & 0.113       \\  
    &  & $\mathcal{L}_{\text{MSE}}$
      & \textbf{0.997} & 0.072 & \textbf{0.942} & \textbf{0.557}  & 0.593   
      & 0.635  & \textbf{0.975}  
      & 0.141   & 0.024    &  0.169    & 0.128       \\

    \midrule
    \multirow{3}{*}{\makecell[l]{Ours \\ \scriptsize (45M)}}  & \multirow{3}{*}{\makecell[c]{\textbf{287ms} \\ }} & $\mathcal{L}_{\text{MSE}}$
      &  0.897 & 0.383 & 0.926 & 0.488 & 0.578    
      & 0.622  & 0.969  
      & 0.151   & 0.074    &  0.061    & 0.087       \\
    &  & $\mathcal{L}_{\text{CMMD}}$
      & \underline{0.949} & 0.271 & \underline{0.931} & 0.517 & \textbf{0.631}    
      & \textbf{0.675}  & \underline{0.973}  
      & \textbf{0.131}   & \textbf{0.004}    &  0.034    & \textbf{0.057}       \\
    &  & $\mathcal{L}_{\substack{\text{CLID} \\ \text{+MSE}}}$
      & 0.902 & 0.405 & 0.929 & 0.506 & \underline{0.627}    
      & \underline{0.672}  & 0.972  
      & \underline{0.140}   & 0.051    &  \underline{0.032}    & 0.068       \\  
    \bottomrule
  \end{tabularx}
  \label{results_experiments}
\end{table*}

\begin{figure}[!t]   
  \centering
  \begin{tabular}{@{}M{2mm}@{\hskip 3pt}M{0.48\columnwidth}@{}M{0.48\columnwidth}@{}}
    \toprule
    & Prediction error & Joint distribution\\
    \midrule
    
    \makecell[c]{\rotatebox[origin=c]{90}{Ours $\mathcal{L}_\text{CMMD}$}} &
    \includegraphics[trim=0 10 0 24,clip,width=\linewidth]{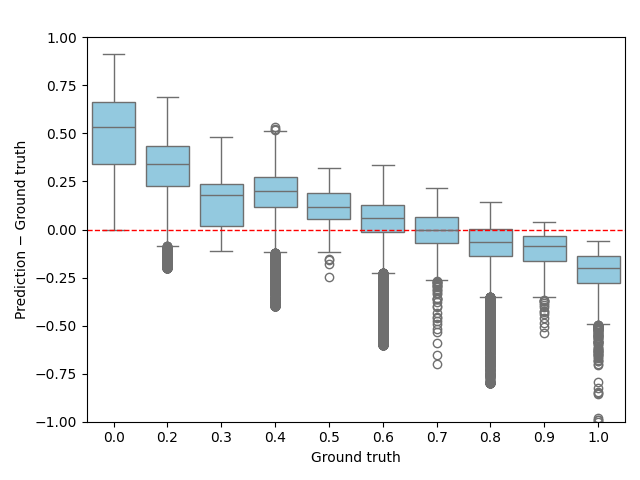} &
    \includegraphics[trim=0 10 17 39,clip,width=\linewidth]{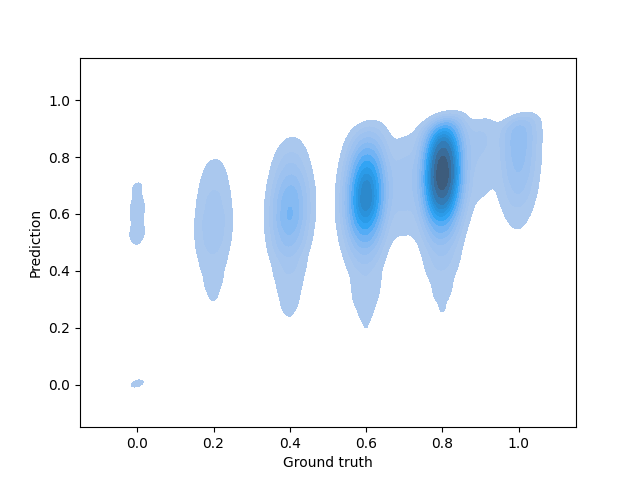} \\

    \midrule
    \makecell[c]{\rotatebox[origin=c]{90}{Qwen}} &
    \includegraphics[trim=0 10 0 24,clip,width=\linewidth]{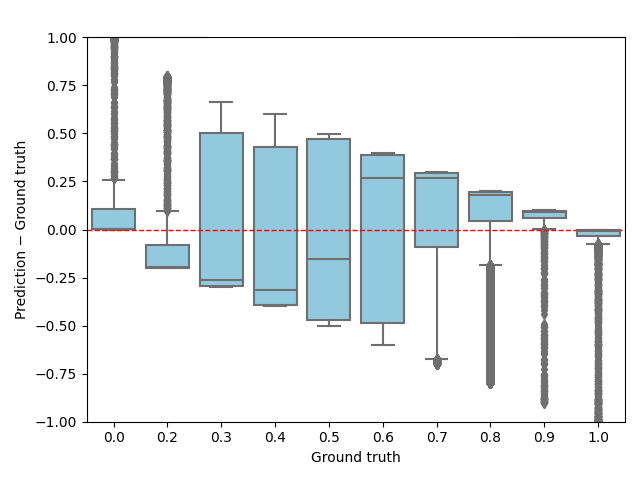} &
    \includegraphics[trim=0 10 17 39,clip,width=\linewidth]{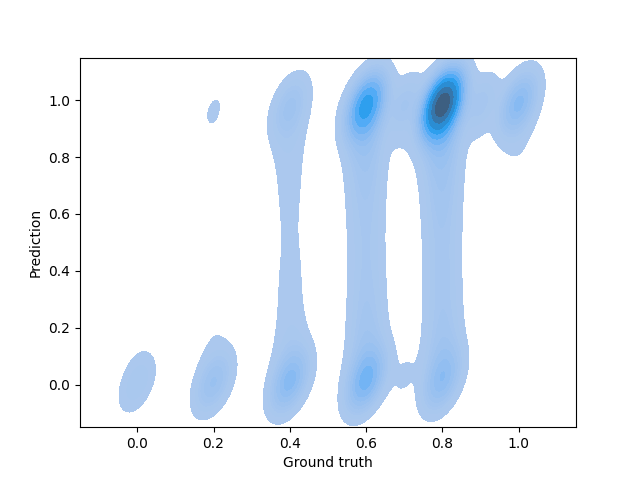} \\
    \makecell[c]{\rotatebox[origin=c]{90}{Qwen $\mathcal{L}_\text{MSE}$}} &
    \includegraphics[trim=0 10 0 24,clip,width=\linewidth]{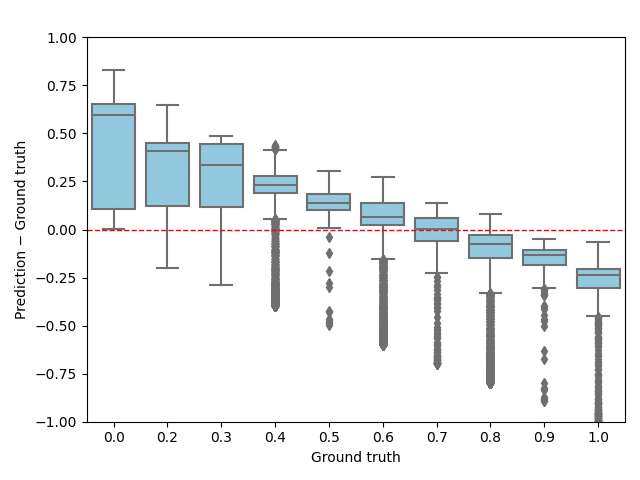} &
    \includegraphics[trim=0 10 17 39,clip,width=\linewidth]{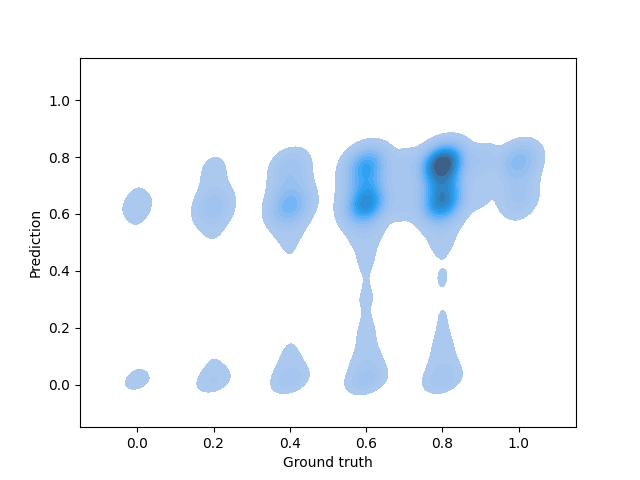} \\

    \midrule
    \makecell[c]{\rotatebox[origin=c]{90}{Gemma}} &
    \includegraphics[trim=0 10 0 24,clip,width=\linewidth]{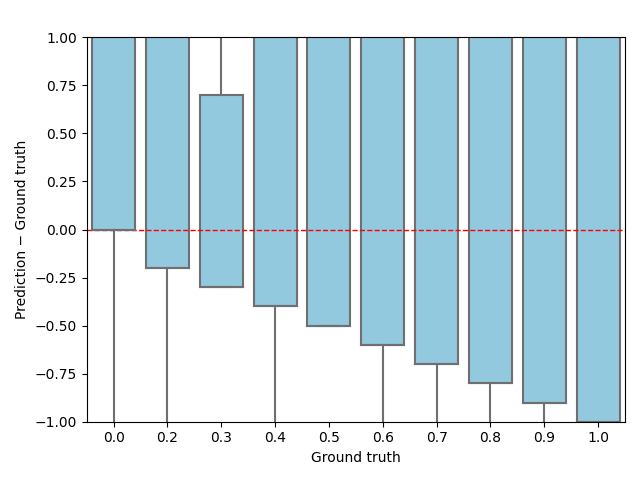} &
    \includegraphics[trim=0 10 17 39,clip,width=\linewidth]{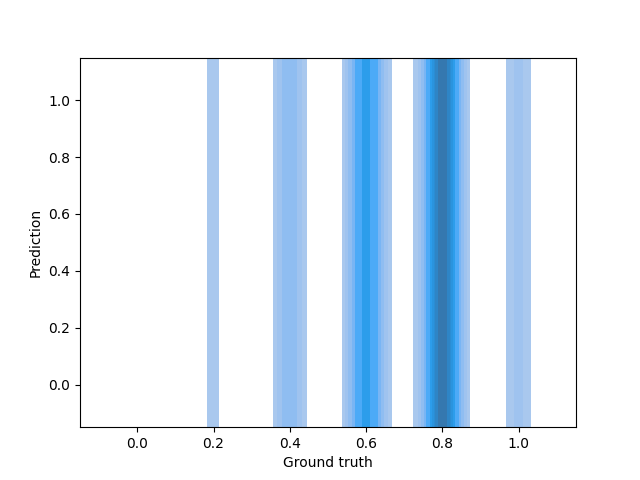} \\
    \makecell[c]{\rotatebox[origin=c]{90}{Gemma $\mathcal{L}_\text{CE}$}} &
    \includegraphics[trim=0 10 0 24,clip,width=\linewidth]{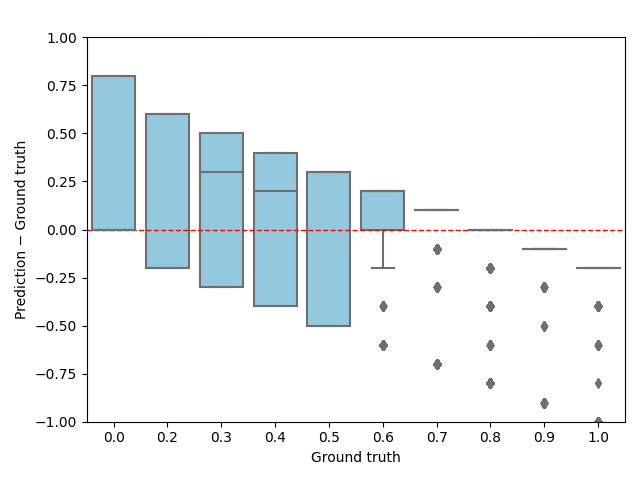} &
    \includegraphics[trim=0 10 17 39,clip,width=\linewidth]{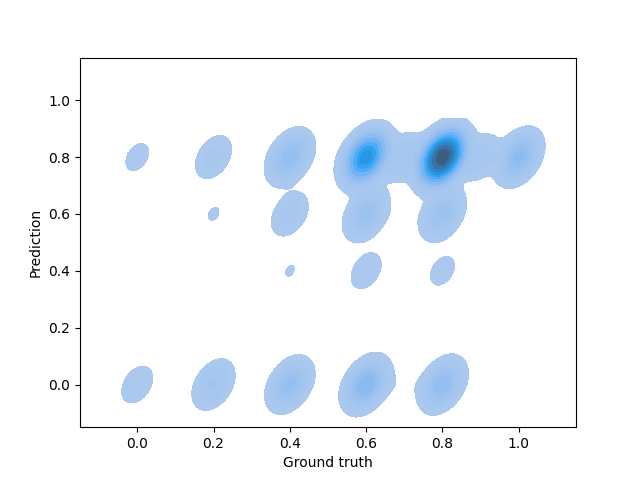} \\
    \makecell[c]{\rotatebox[origin=c]{90}{Gemma $\mathcal{L}_\text{MSE}$}} &
    \includegraphics[trim=0 0 0 24,clip,width=\linewidth]{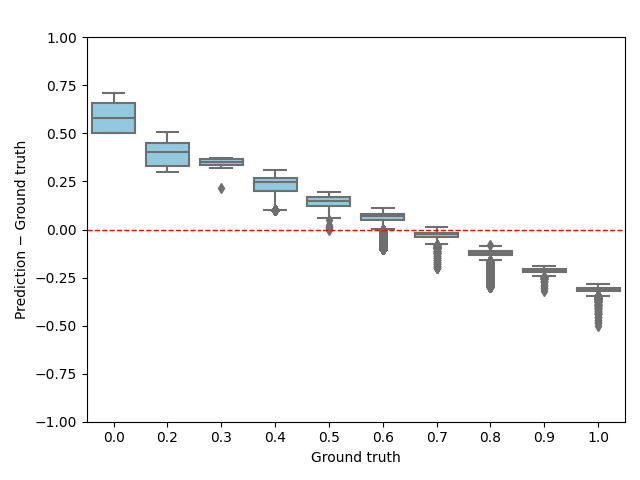} &
    \includegraphics[trim=0 10 17 39,clip,width=\linewidth]{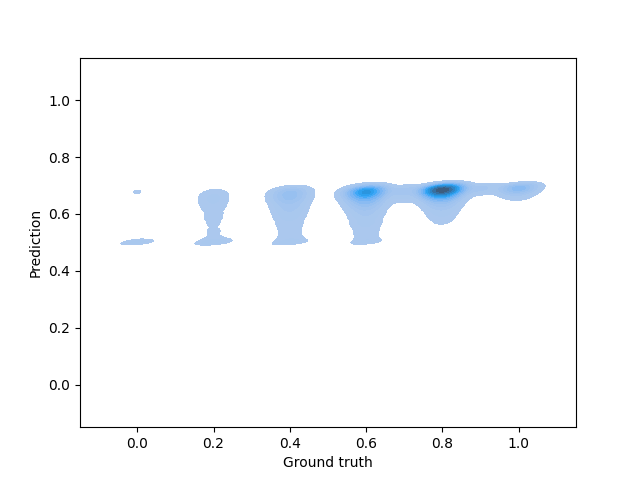} \\

    \bottomrule
  \end{tabular}
    \caption{Box plots of prediction errors grouped by discrete ground-truth scores (left) and joint distribution between ground-truth scores and model predictions (right) for each method.
    }
  \label{fig:dist_eval}
\end{figure}

Table~\ref{results_experiments} presents evaluation results on the test set (Section~\ref{dataset_description}), using the metrics from Section~\ref{sec:evaluation-metrics}. It compares all models from Section~\ref{models_description}, in both zero-shot and fine-tuned settings, as described in Section~\ref{training_description}.

Figure~\ref{fig:dist_eval} complements the evaluation with two plots per model: a box plot (left) showing prediction errors across ground-truth scores, and a kernel density estimates (right) of the joint distribution between predicted and true scores. Each row corresponds to a different model. \\

In terms of relevance evaluation, we observe a trade-off between discarding non-relevant candidates and preserving relevant ones. Qwen3 performs well in zero-shot, particularly on specificity and R-Precision (R-P), but fine-tuning appears to harm its generalization. In contrast, regression-based fine-tuning significantly boosts Gemma3's performance, especially in recall, R-P, and $\Bar{R}$-$\mathcal{O}$. Our model is competitive, being second on key relevance metrics and performing well in specificity.

Looking at ranking metrics such as mean average precision (mAP), mean reciprocal rank (MRR), and normalized discounted cumulative gain (NDCG), our method achieves consistently strong performance. When trained with $\mathcal{L}_\text{CMMD}$, it consistently achieves top performance across all metrics. Gemma3 slightly outperforms on NDCG (0.975 vs. 0.973), while our model with CLID loss is a close second in overall ranking quality.

Calibration analysis based on Figure~\ref{fig:dist_eval} reveals that Qwen3 often produces extreme scores (close to $0$ or $1$), suggesting poor calibration despite good binary discrimination. Fine-tuning helps mitigate this but does not fully resolve the issue. Gemma3 in zero-shot generates a wide range of hallucinated scores, which are corrected with fine-tuning. However, next-token tuning introduces discretization that appears misaligned with Gemini 2.0's scoring, and regression reduces the expressiveness of the scores. Overall on this aspect, Our method provides the best alignment with Gemini 2.0 scores, both visually and based on evaluation metrics.

In terms of efficiency, our model is highly scalable, processing 1,000 profile-brief pairs in under one minute, or just 287 milliseconds when using precomputed profile embeddings. In comparison, Qwen3 requires 7 minutes and Gemma3 about 13 minutes. This substantial speed advantage makes our approach more practical for real-time or large-scale deployments, including CPU-only environments. \\

To assess potential bias, we conducted a simple gender-based fairness analysis. The test set was split by gender declared by freelancers when creating their profiles, and recall was computed for each gender group. The difference in recall between women and men was 0.005 using historical labels and 0.010 using Gemini 2.0 scores, suggesting our model does not discriminate relevant freelancers based on their gender. We acknowledge this is a preliminary assessment and that a deeper fairness analysis~\cite{bied2023fairness} would be beneficial. \\

Lastly, our model demonstrates good robustness to out-of-distribution samples (Appendix~\ref{sec:out-of-distrib}), while retaining semantic alignment from the frozen backbone (Appendix~\ref{sec:robust-language}). However, performance on synthetic average-match cases indicates room for improvement, particularly in distinguishing non-relevant profiles.
\section{Conclusion}
\label{sec:conclusion}

This paper presents a lightweight model for long-context multilingual reranking of project–freelancer pairs, leveraging a distillation framework to produce semantically calibrated and interpretable scores. Our two-step architecture, comprising two encoding branch followed by a comparison block, outperforms both zero-shot and fine-tuned baselines on relevance, ranking, and calibration metrics, demonstrating its effectiveness for skill relevance assessment. Furthermore, the proposed utterance-based encoding strategy significantly reduces computational complexity, enabling efficient processing of long documents. The ability to precompute freelancer profile embeddings further supports low-latency inference, making the model well-suited for real-time deployment in production environments.

Future work will focus on refining key components of the distillation framework. In particular, careful attention should be paid to the construction of training and evaluation datasets. A dedicated test set derived from historical data with high-quality labels is essential to better evaluate calibration and interpretability. This may require debiasing the data and conducting a label annotation campaign to introduce finer-grained, calibrated labels.

In addition, deeper analysis of potential biases, especially under production conditions, is essential. We also plan to extend the comparison of the teacher’s scores with expert judgments to better assess its own calibration. Since our supervision relies on synthetic labels, we must remain cautious about inherited biases and explore strategies to monitor and mitigate them. Long-term robustness will require handling potential drift of the teacher model, for example through periodic re-calibration or self-distillation.
Finally, future research should investigate how this model can be integrated into downstream systems, for example as a feature within existing ranking algorithms or as a tool to improve transparency and interpretability in user-facing applications. Indeed, an important next step will be controlled online experiments to assess business impact.
\section*{Declaration on Generative AI}

During the preparation of this work, the authors used  \textit{chat-GPT 3.5} and \textit{Gemini-2.5-flash}  for grammar and spelling check. After using these tools, the authors reviewed and edited the content as needed and take full responsibility for the publication’s content.

{\small \bibliography{biblio}}

\newpage
\appendix
\section{Statistics description of utterances interactions}
\label{sec:stats-interaction}
Due to the varying length of the similarity distributions $\mathcal{S}_{p,f}$ and $\mathcal{S}_{f,p}$, we extract a set of descriptive statistics from each distribution, in order to get fixed size features more suitable for scoring. These descriptive statistics are defined as the following for $\mathcal{S}_{p,f}$ : 
\begin{equation}
    \small
    \operatorname{desc}(\mathcal{S}_{p,f}) = 
    \begin{cases}
        \displaystyle\min_{\rho \in \mathcal{S}_{p,f}}(\rho)
        & \textbf{Minimum} \\[0.5em]
        \displaystyle\max_{\rho \in \mathcal{S}_{p,f}}(\rho)
        & \textbf{Maximum} \\[1em]
        
        \overline{\vphantom{S^{s}}\mathcal{S}}_{p,f} = \displaystyle\sum_{\rho \in \mathcal{S}_{p,f}} \frac{\rho}{|\mathcal{S}_{p,f}|}
        & \textbf{Mean} \\[1em]
        
        \sigma_{p,f} = \displaystyle\sqrt{\sum_{\rho \in \mathcal{S}_{p,f}} \frac{(\rho - \overline{\vphantom{S^{s}}\mathcal{S}}_{p,f})^2}{|\mathcal{S}_{p,f}|}}
        & \raisebox{0.75em}{\begin{array}[t]{@{}l} \textbf{Standard}\\\textbf{deviation} \end{array}} \\[1.5em]

        \displaystyle\frac{1}{|\mathcal{S}_{p,f}|}\sum_{\rho \in \mathcal{S}_{p,f}} 
        \left(\frac
        {\rho - \overline{\vphantom{S^{s}}\mathcal{S}}_{p,f}}
        {\sigma_{p,f}} \right) ^3
        & \textbf{Skewness} \\[1em]
        
        \displaystyle\frac{1}{|\mathcal{S}_{p,f}|}\sum_{\rho \in \mathcal{S}_{p,f}} 
        \left(\frac
        {\rho - \overline{\vphantom{S^{s}}\mathcal{S}}_{p,f}}
        {\sigma_{p,f}} \right) ^4
        & \textbf{Kurtosis} \\
    \end{cases}
\end{equation}
The same statistics are computed based on $\mathcal{S}_{f,p}$ to obtain $\operatorname{desc}(\mathcal{S}_{f,p})$.

\section{Robustness: Impact of Brief Language}
\label{sec:robust-language}
\begin{table*}[ht]
  \caption{Model performance across brief languages. \textcolor{darkred}{\textbf{Red}} denotes the worst value; \textcolor{darkgreen}{\textbf{Green}} highlights the best.}
  \centering
  \small
  \setlength{\tabcolsep}{8pt} 
  \begin{tabularx}{\linewidth}{@{}l@{\hskip 3pt}Y *{5}{Y} *{2}{Y} *{4}{Y}@{}}
    \toprule
    \multirow{2}{*}{\makecell[l]{\textbf{Brief} \\ \textbf{Language}}} & \multirow{2}{*}{\textbf{Support}}
    & \multicolumn{5}{c}{\textbf{Relevancy} $\scriptstyle(s_t > 0.5)$}
    & \multicolumn{2}{c}{\textbf{Ranking}}
    & \multicolumn{4}{c}{\textbf{Calibration} (distances)} \\
    \cmidrule(lr){3-7} 
    \cmidrule(lr){8-9} 
    \cmidrule(lr){10-13}
     & 
    & Rec. & Spec. & R-P & $\Bar{\text{R}}$-$\mathcal{O}$ & mAP 
    & MRR & NDCG
    & MAE & $\Delta_{mean}$ & $\Delta_{IQR}$ & $W_1$ \\
    
    \midrule
    French & 6277
    & 0.951 & 0.264 & 0.936 & 0.513 & 0.632
    & 0.677 & 0.974
    & 0.129 & 0.005 & \textcolor{darkgreen}{\textbf{0.033}} & \textcolor{darkgreen}{\textbf{0.056}} \\
    
    English & 770
    & 0.928 & 0.315 & \textcolor{darkred}{\textbf{0.890}} & 0.543 & 0.610
    & 0.644 & \textcolor{darkred}{\textbf{0.962}}
    & \textcolor{darkred}{\textbf{0.146}} & \textcolor{darkgreen}{\textbf{0.003}} & 0.045 & \textcolor{darkred}{\textbf{0.068}} \\
    
    Spanish & 605
    &0.950 & 0.297 & 0.924 & \textcolor{darkred}{\textbf{0.503}} & 0.668
    & \textcolor{darkgreen}{\textbf{0.705}} & 0.974
    & 0.136 & \textcolor{darkgreen}{\textbf{0.003}} & 0.044 & 0.063 \\
    
    German & 357
    & \textcolor{darkgreen}{\textbf{0.959}} & \textcolor{darkred}{\textbf{0.171}} & 0.932 & \textcolor{darkgreen}{\textbf{0.552}} & \textcolor{darkred}{\textbf{0.583}}
    & \textcolor{darkred}{\textbf{0.627}} & 0.971
    & \textcolor{darkgreen}{\textbf{0.126}} & 0.022 & \textcolor{darkred}{\textbf{0.054}} & 0.063 \\
    
    Dutch & 30
    & \textcolor{darkred}{\textbf{0.925}} & \textcolor{darkgreen}{\textbf{0.421}} & \textcolor{darkgreen}{\textbf{0.946}} & 0.513 & \textcolor{darkgreen}{\textbf{0.690}}
    & 0.690 & \textcolor{darkgreen}{\textbf{0.976}}
    & 0.130 & \textcolor{darkred}{\textbf{0.023}} & 0.047 & \textcolor{darkgreen}{\textbf{0.056}} \\
    
    \bottomrule
  \end{tabularx}
  \label{tab:language-results}
\end{table*}

\begin{table*}[ht]
  \caption{Impact of out-of-distribution examples on model robustness. The test set is enriched with synthetic \textit{average} or \textit{unsuitable} matches. \textcolor{darkred}{\textbf{Red}} highlights show degraded performance, \textcolor{darkgreen}{\textbf{green}} indicates improvement or consistency.}
  \centering
  \small
  \setlength{\tabcolsep}{8pt} 
  \begin{tabularx}{\linewidth}{@{}l *{5}{Y} *{2}{Y} *{4}{Y}@{}}
    \toprule
    \multirow{2}{*}{\textbf{Test data}}
    & \multicolumn{5}{c}{\textbf{Relevancy} $\scriptstyle(s_t > 0.5)$}
    & \multicolumn{2}{c}{\textbf{Ranking}}
    & \multicolumn{4}{c}{\textbf{Calibration} (distances)} \\
    \cmidrule(lr){2-6} 
    \cmidrule(lr){7-8} 
    \cmidrule(lr){9-12}
     & 
     Rec. & Spec. & R-P & $\Bar{\text{R}}$-$\mathcal{O}$ & mAP 
    & MRR & NDCG
    & MAE & $\Delta_{mean}$ & $\Delta_{IQR}$ & $W_1$ \\
    
    \midrule
    historical interactions $\spadesuit$
      & 0.949 & 0.271 & 0.931 & 0.517 & 0.631  
      & 0.675  & 0.973
      & 0.131   & 0.004   &  0.034    & 0.057      \\

    $\spadesuit$ + \textit{average}
      & \textcolor{darkred}{\textbf{0.885}} & \textcolor{darkgreen}{\textbf{0.363}} & \textcolor{darkred}{\textbf{0.790}} & \textcolor{darkgreen}{\textbf{0.529}} & \textcolor{darkred}{\textbf{0.548}}
      & 0.604  & 0.963
      & 0.132   & 0.026   &  0.026    & 0.069      \\

    $\spadesuit$ + \textit{unsuitable}
      & \textcolor{darkgreen}{\textbf{0.949}} & \textcolor{darkgreen}{\textbf{0.861}} & 0.886 & \textcolor{darkgreen}{\textbf{0.938}} & 0.624   
      & 0.669  & \textcolor{darkred}{\textbf{0.957}}
      & \textcolor{darkgreen}{\textbf{0.103}}   & 0.033   &  \textcolor{darkred}{\textbf{0.096}}    & 0.063      \\
    \bottomrule
  \end{tabularx}
  \label{tab:synthetic-results}
\end{table*}

Table~\ref{tab:language-results} reports the performance of our model (Section~\ref{proposed_architecture}), trained with the $\mathcal{L}_\text{CMMD}$ loss (Eq.~\ref{eq:ours-margin-mse}) and evaluated using the metrics from Section~\ref{sec:evaluation-metrics}, on test set splits by brief language. Rows are ordered by language frequency. This analysis assesses the model’s multilingual robustness, despite relying solely on a multilingual backbone without additional language-specific training.\\

French, being the most represented language, yields “average” performance across most metrics. This confirms that the model does not overfit to the dominant language in a way that degrades generalisation.\\

Interestingly, Dutch achieves the strongest results across most metrics, despite having the smallest support. This can be explained by the recent introduction of Dutch on the platform, which involved more manual curation, human refinement, and assisted onboarding. We hypothetize that These factors have led to cleaner training signals.\\

German shows excellent recall but poor specificity and ranking metrics, suggesting that the model tends to overestimate relevance in that language. This may reflect weaker signal quality or domain mismatch.\\

English underperforms on multiple fronts, particularly in calibration and ranking. As English is used in many different regions and contexts, the briefs likely exhibit higher lexical and stylistic variability, which may introduce noise during training.\\

Overall, the model shows strong generalization across languages, including underrepresented ones. This highlights the robustness of the approach, even in the absence of multilingual-specific objectives or balancing strategies. However, performance gaps observed in certain languages, particularly German and English, suggest that enhancing the quality of training signals in these languages could further improve results.

\section{Robustness: Impact of Out-of-Distribution Samples}
\label{sec:out-of-distrib}
Table~\ref{tab:synthetic-results} reports the performance evaluation results for our proposed model (Section~\ref{proposed_architecture}), trained with the $\mathcal{L}_\text{CMMD}$ loss (Eq.~\ref{eq:ours-margin-mse}) and evaluated using the metrics defined in Section~\ref{sec:evaluation-metrics}. The original test set (first row) is enriched with out-of-distribution samples: \textit{average} matches in the second row, and \textit{unsuitable} matches in the last row. The synthetic \textit{average} and \textit{unsuitable} samples are constructed using the same heuristics as those used during training. \\

The borderline \textit{average} matches pose a challenge: the model's recall and precision (R-P) decrease, indicating difficulty in confidently labeling them as relevant. However, specificity and $\Bar{R}-\mathcal{O}$ improve, meaning the model successfully avoids over-recommending these borderline candidates. This behavior leads to a slightly worse mAP, likely because some relevant items are misclassified as irrelevant. \\

The model remains robust to "unsuitable" matches, with unchanged recall and significant gains in specificity. Calibration metrics are stable or slightly improved, indicating good semantic separation. However, NDCG shows a small drop, suggesting some unsuitable candidates may still rank above borderline ones, an aspect for further exploration. \\

In conclusion, the model demonstrates strong robustness to unknown interactions. It conservatively handles average cases and reliably down ranks unsuitable ones. This behavior is desirable in production, where pushing weak or irrelevant recommendations should be avoided, supporting its deployment in real-world scenarios. Nonetheless, additional investigation is needed to mitigate undesirable behaviors, such as the misclassification of relevant candidates or the overranking of unsuitable ones.

\end{document}